\documentclass[journal,11pt,onecolumn]{IEEEtran}
\ifCLASSINFOpdf
\else
\fi

\usepackage{graphicx}
\usepackage{verbatim}
\usepackage{latexsym}
\usepackage{setspace}

\usepackage{bibunits}
\usepackage{hhline}

\usepackage{mathtools}

\usepackage{amsmath}
\newcommand\numeq[1]%
  {\stackrel{\scriptscriptstyle \mkern-1.5mu#1 \mkern-1.5mu}{\eqdef}}

\DeclarePairedDelimiter{\norm}{\lVert}{\rVert}
\usepackage[T1]{fontenc}
\usepackage{mathpazo}

\usepackage{extarrows}
\newcommand{\eqdef}{\xlongequal{\text{$\norm{W_j}=1$}}}%

\newcommand\footnoteref[1]{\protected@xdef\@thefnmark{\ref{#1}}\@footnotemark}

\usepackage{times}
\usepackage{epsfig}
\usepackage{graphics}
\usepackage{amsmath}
\usepackage{amssymb}
\usepackage{caption} 
\usepackage[export]{adjustbox}
\usepackage{comment}
\usepackage[colorinlistoftodos]{todonotes}
\usepackage{cite}

\usepackage{tabularx}

\usepackage{subfigure}
\usepackage{booktabs}
\usepackage{rotating}
\usepackage{multirow}
\usepackage{color}
\usepackage{xspace}

\usepackage{enumitem}

\usepackage{hyperref}
\hypersetup{
    colorlinks=true,
    linkcolor=blue,
    filecolor=magenta,      
    urlcolor=red,
}
\begin{document}
%
\title{Affect Analysis in-the-wild: Valence-Arousal, Expressions, Action Units and a Unified Framework}
%
%
%

\author{Dimitrios~Kollias,~\IEEEmembership{Member,~IEEE,}
        Stefanos~Zafeiriou
\thanks{D. Kollias is with the School of Computing and Mathematical Sciences, University of Greenwich, London, UK, e-mail: D.Kollias@greenwich.ac.uk}
\thanks{S. Zafeiriou is with the Department of Computing, Imperial College London, UK, e-mail: s.zafeiriou@imperial.ac.uk}
}

%
%

\markboth{}%
{D.Kollias \MakeLowercase{\textit{et al.}}: Affect Analysis in-the-wild: Valence-Arousal, Expressions, Action Units and a Unified Framework}
%



\maketitle

\begin{abstract}
Affect recognition based on subjects' facial expressions has been a topic of major research in the attempt to generate machines that can understand the way subjects feel, act and react. 
In the past, due
to the unavailability of large amounts of data captured in real-life situations, research has mainly focused on controlled environments.
However, recently, social media and platforms have been widely used. Moreover, deep learning has
emerged as a means to solve visual analysis and recognition problems. 
This paper exploits these advances and
presents significant contributions for affect analysis and recognition in-the-wild.
Affect analysis and recognition can be seen as a dual knowledge generation problem, involving: i) creation of new, large and rich in-the-wild databases  and ii) design and training of novel deep neural architectures that are able to analyse affect over these databases and to successfully generalise their performance on other datasets.
The paper focuses on large in-the-wild databases, i.e., Aff-Wild and Aff-Wild2 and presents the design of two classes of deep neural networks trained with these databases. The first class refers to uni-task affect recognition, focusing on prediction of the valence and arousal dimensional variables. The second class refers to     
estimation of all main behavior tasks, i.e. valence-arousal prediction; categorical emotion classification in seven basic facial expressions; facial Action Unit detection.
A novel multi-task and 
holistic framework is presented which is able to jointly learn 
and effectively generalize and perform affect recognition over all existing in-the-wild databases. Large experimental studies illustrate the achieved performance improvement over the existing state-of-the-art in affect recognition.

\end{abstract}


%
\IEEEpeerreviewmaketitle

\section{Introduction}
%
%
%
%
\IEEEPARstart{T}{his} paper presents recent developments and research directions in  affective behavior analysis in-the-wild, which is a major targeted characteristic of human computer interaction systems in real life applications.  Such systems,  machines and robots, should be able to  automatically sense and interpret facial and audio-visual signals relevant to emotions, appraisals and intentions; thus, being able to interact in a 'human-centered' and engaging manner with people, as their digital assistants in the home,  work, operational or industrial environment.

Through human affect recognition, the reactions of the machine, or robot, will be consistent with people's expectations and emotions; their verbal and non-verbal interactions will be positively received by humans. Moreover, this interaction should not be dependent on the respective context, nor the human's age, sex, ethnicity, educational level, profession, or social position. As a consequence, the development of intelligent systems able to analyze human behavior in-the-wild can contribute to generation of trust, understanding and closeness between humans and machines in real life environments. 

In this paper we mainly focus on facial affect analysis, which constitutes a difficult problem, because emotion patterns in faces are complex, time varying, user and context dependent. We deal with all three main emotion models, i.e., facial expressions and categorical affect, facial action units and dimensional affect representations.

Representing human emotions has been a basic topic of research in psychology. The most frequently used emotion representation is the categorical one, including the seven basic categories, i.e., Anger, Disgust, Fear, Happiness, Sadness, Surprise and Neutral \cite{ekman2002facial}. Discrete emotion representation can also be described in terms of the Facial Action Coding System (FACS) model, in which all possible facial actions are described in terms of Action Units (AUs) \cite{ekman2002facial}. Finally, the dimensional model of affect \cite{whissel1989dictionary,russell1978evidence} has been proposed as a means to distinguish between subtly different displays of affect and encode small changes in the intensity of each emotion on a continuous scale. The 2-D Valence and Arousal Space (VA-Space) is the most usual dimensional emotion representation; valence shows how positive or negative an emotional state is, whilst arousal shows how passive or active it is.

Emotion recognition is a very important topic in human computer interaction \cite{kollias2015interweaving}.   Furthermore, since real-world settings entail uncontrolled conditions, where subjects operate in a diversity of contexts and environments, systems that perform automatic analysis of human behavior and emotion recognition should be robust to video recording conditions, diversity of contexts and timing of display.
These goals are scientifically and technically challenging.
We wouldn't be able to create emotion recognition systems \cite{kollias2016line} that work on real life situations if we did not have large emotion databases that simulate behaviors in-the-wild. 

Research in face perception and emotion analysis has, therefore, been greatly assisted by the creation of databases \cite{tagaris2018machine} of images and video sequences annotated in terms of persons' behaviors, their facial expressions and underlying emotions.  To this end, some in-the-wild datasets have been generated  and used for recognition of facial expressions, for detection of facial action units and for estimation of valence and arousal. However:  i) their size is small,  ii) they are not audiovisual, iii)  only  a  small  part  of them is  manually  annotated,  iv)  they  contain  a  small  number of subjects, v) they are not annotated for all the tasks.
In this paper, we focus and use a class of new in-the-wild databases, Aff-Wild \cite{zafeiriou2017aff}\cite{zafeiriou} and Aff-Wild2 \cite{kollias2019expression,kollias2020analysing} that overcome the above limitations.

Current state-of-the-art affect recognition methods are generally based on Deep Neural Networks (DNNs) \cite{tagaris2017assessment}
that are able to analyze large amounts of data and recognise persons' emotional states. In this paper we present new DNN systems that can be trained with the above-mentioned large in-the-wild databases and  improve the state-of-the-art in affect recognition, considering either the dimensional, categorical, or AU based emotion models. 

Furthermore, we develop multitask architectures that can jointly learn to recognize all three emotion representations in-the-wild. We present  a novel holistic approach that achieves excellent performance, by coupling all three recognition tasks during training \cite{kollias2019face}. To achieve this, we utilise all publicly available datasets (including over 5 million images) that study facial behavior tasks in-the-wild. 

The rest of the paper is organized as follows. Section \ref{models_affect} briefly describes the three models of affect that are adopted and used for the development of affect recognition methods. Section \ref{dbs} presents the existing databases and their properties, focusing on the Aff-Wild and Aff-Wild2 databases. The  signal pr-processing steps and the performance evaluation criteria are described in Section \ref{pre-processing}. Section \ref{uni-task}
examines uni-task affect analysis, focusing on dimensional affect; it presents single-component and multiple-component architectures which improve the state-of-the-art in dimensional affect recognition. The unified affect analysis methodology, over all three emotion models, is described in Section \ref{mt-task}. It first presents the multitask affect analysis approach and then the unifying  holistic  approach. Section \ref{conclusion} presents the conclusions and further prospects of the presented approaches.

\section{Models of Affect}\label{models_affect}

\subsection{Categorical Affect}

 Ekman defined the six basic emotions, i.e., Anger, Disgust, Fear, Happiness, Sadness, Surprise and the neutral state, based on a cross-culture study\cite{ekman2002facial}, which indicated that humans perceive certain basic emotions in the same way regardless of culture. Nevertheless, advanced research on neuroscience and psychology argued that the model of six basic emotions are culture-specific and not universal. Additionally, the affect model based on basic emotions is limited in the ability to represent the complexity and subtlety of our daily affective displays. Despite these findings, the categorical model that describes emotions in terms of discrete basic emotions is still the most popular perspective for Facial Expression Recognition (FER), due to its pioneering
investigations along with the direct and intuitive definition of facial expressions.




\subsection{Action Units}

Detection of Facial Action Units (AUs) has also attained large attention. The Facial Action Coding System (FACS) \cite{ekman2002facial} provides a standardised taxonomy of facial muscles' movements and has been widely adopted as a common standard towards systematically categorising physical manifestation of complex facial expressions. Since any facial expression can be represented as a combination of action units, they constitute a natural physiological basis for face analysis. The existence of such a basis is a rare boon for computer vision, as it allows focusing on the essential atoms of the problem and, by virtue of their exponentially large possible combinations, opens the door for studying a wide range of applications beyond prototypical emotion classification. Consequently, in the last years, there has been a shift of related research towards the detection of action units. The presence of action units is typically brief and unconscious, and their detection requires analyzing subtle appearance changes in the human face. Furthermore, action units do not appear in isolation, but as elemental units of facial expressions, and hence some AUs co-occur frequently, while others are mutually exclusive. Fig. \ref{whicsel} shows the most common action units and the corresponding facial action movement that defines them. 

\subsection{Dimensional Affect}

\begin{figure*}
\centering
\adjincludegraphics[width=1\linewidth]{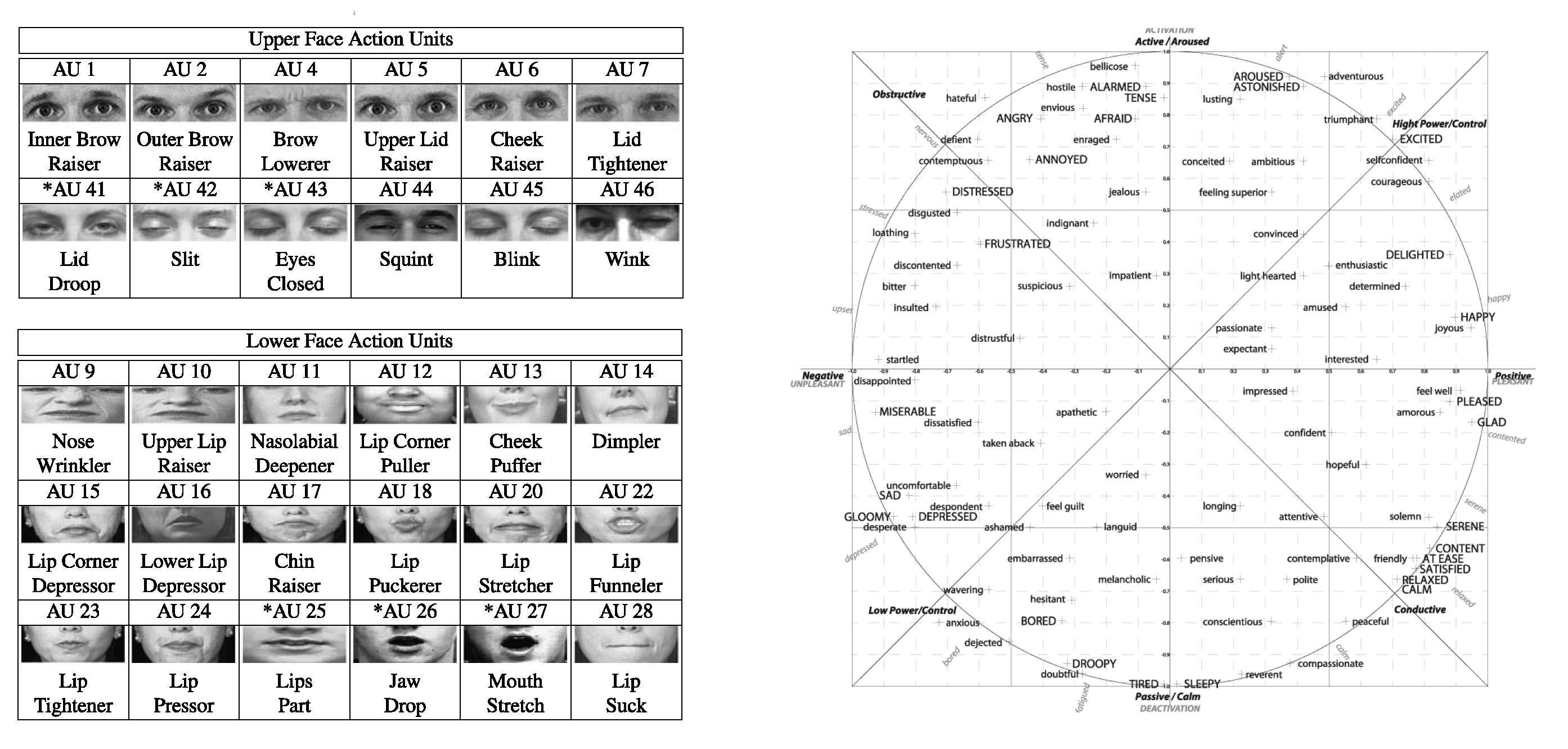} 
\caption{The 2D Valence-Arousal Space \& Some facial Action Units}
\label{whicsel}
\end{figure*}

The dimensional model of affect, which is appropriate to represent not only extreme, but also subtle emotions appearing in everyday human-computer interactions, has also attracted significant attention over the last years.  According to the dimensional approach \cite{russell1978evidence} \cite{whissel1989dictionary}, affective behavior is described by a number of latent continuous dimensions. The most commonly used dimensions include valence (indicating how positive or negative an emotional state is) and arousal (measuring the power of emotion activation). Valence and arousal readily relate to specific functions of regions of the brain; the parietal region of the right hemisphere appears to play a special role in the mediation of arousal, whereas the frontal regions appear to play a special role in emotional valence. Fig. \ref{whicsel} shows the 2D Valence-Arousal Space, in which the horizontal axis denotes valence that ranges from very positive to very negative and the vertical one denotes arousal that ranges from very active to very passive. 

\section{Existing Datasets with Affect Annotation}\label{dbs}

Current research in automatic analysis of facial affect aims at developing systems, such as robots and virtual humans, that will interact with humans in a naturalistic way under real-world settings. To this end, such systems should automatically sense and interpret facial signals relevant to emotions, appraisals and intentions.  Moreover, since real-world settings entail uncontrolled conditions, where subjects operate in a diversity of contexts and environments, systems that perform automatic analysis of human behavior should be robust to video recording conditions, the diversity of contexts and the timing of display.

For the past twenty years research in automatic analysis of facial behavior was mainly limited to posed behaviors which were captured in highly controlled recording conditions. 
Nevertheless, it is now accepted by the community that facial expressions of naturalistic behaviors can be radically different from posed ones. Hence, efforts have been made in order to collect subjects displaying naturalistic behavior. Examples include the DISFA \cite{mavadati2013disfa}, BP4D-Spontaneous (BP4DS) \cite{zhang14bp4d}, BP4D+ \cite{zheng2018multimodal} databases.

However, with the development of large and diverse datasets in the field of computer vision (and the accompanying performance gains), it has become apparent that the diversity of human participants and spontaneous expressions  have  to  become the  prerogatives  in  deployment  of  the  affective  computing  models  in practice. Hence, it is now widely accepted, in both the computer vision and machine learning communities, that progress in a particular application domain is significantly catalyzed when  large datasets are collected in unconstrained conditions (also referred as "in-the-wild" data). Therefore, facial analysis could not only focus on spontaneous behaviors, but also on behaviors captured in unconstrained conditions. 

Some datasets with in-the-wild settings have been recently collected to study: i) facial expression analysis, such as the static AffectNet \cite{mollahosseini2017affectnet} and the static RAF-DB \cite{li2017reliable}; ii) facial action units, such as the static EmotioNet \cite{benitez2017emotionet}; and iii) continuous emotions of valence and arousal in-the-wild, such as
the audiovisual SEWA \cite{kossaifi2019sewa}, the
audiovisual OMG-Emotion  \cite{barros2018omg},  the static AffectNet \cite{mollahosseini2017affectnet} and the static AFEW-VA \cite{kossaifi2017afew} datasets. Let us note that the term 'static' means that the dataset contains only (static) images, neither video nor audio.
Table \ref{fer_dbs} shows some in-the-wild, databases that exist in literature (and are being utilised in our experiments in the next Sections) and are annotated in terms of either facial expressions, or action units, or valence-arousal.
The in-the-wild databases are captured under different illumination conditions in uncluttered backgrounds and contexts, in which people have different head poses and there exist occlusions in the facial area.

\begin{table*}
\centering
\caption{Existing Databases along with their properties;  'static' means images, 'dynamic' means image sequences (video without audio), 'A/V' means audiovisual sequences  (video with audio) }
\label{fer_dbs}
\scalebox{0.82}{
\begin{tabular}{ |c||c|c|c|c|c|  }
\hline 
 DBs & DB Type & Model of Affect & Condition & DB Size    \\
\hhline{|=|=|=|=|=|} 
\hline
RAF-DB\cite{li2017reliable} & static & \begin{tabular}{@{}c@{}} 6 Basic,  Neutral \& 11 Compound \end{tabular}  & in-the-wild  & \begin{tabular}{@{}c@{}} 15,339 \& 3,954 \end{tabular}     \\
\hline
AffectNet\cite{mollahosseini2017affectnet} & static & \begin{tabular}{@{}c@{}} 6 Basic, Neutral + Contempt \\ valence-arousal \end{tabular} & in-the-wild  & \begin{tabular}{@{}c@{}}291,651 manual \& 400,000 automatic annotations \\ 325,000 manual \& 460,300 automatic  annotations  \end{tabular}   \\
\hline
DISFA \cite{mavadati2013disfa}  & dynamic &  \begin{tabular}{@{}c@{}} 12 action units  \end{tabular} & controlled  & \begin{tabular}{@{}c@{}} 54 videos: 261,630 frames \end{tabular} \\
\hline
BP4DS \cite{zhang14bp4d} & dynamic &  \begin{tabular}{@{}c@{}} 27 action units \end{tabular} & controlled  & \begin{tabular}{@{}c@{}} 1,640 videos: 222,573 frames \end{tabular} \\
\hline
BP4D+ \cite{zheng2018multimodal} & dynamic &  \begin{tabular}{@{}c@{}} 34 action units  \end{tabular} & controlled  & \begin{tabular}{@{}c@{}} 5,463 videos: 967,570 frames \end{tabular}\\
\hline
EmotioNet \cite{benitez2017emotionet} & static &  \begin{tabular}{@{}c@{}} 11 action units  \end{tabular} & in-the-wild  & \begin{tabular}{@{}c@{}} 50,000 manual \& 950,000 automatic annotations  \end{tabular} \\
\hline
AFEW-VA \cite{kossaifi2017afew} & dynamic  &  \begin{tabular}{@{}c@{}} valence-arousal \end{tabular} & in-the-wild  & \begin{tabular}{@{}c@{}} 600 videos: 30,050 frames \end{tabular} \\
\hline
SEWA \cite{kossaifi2019sewa} & A/V  &  \begin{tabular}{@{}c@{}} valence-arousal \end{tabular} & in-the-wild  & \begin{tabular}{@{}c@{}} 538 videos  \end{tabular}  \\
\hline
OMG-Emotion \cite{barros2018omg} & A/V  &  \begin{tabular}{@{}c@{}} valence-arousal \end{tabular} & in-the-wild  & \begin{tabular}{@{}c@{}} 495 videos: 5,288 utterances  \end{tabular} \\
\hline
\end{tabular}
}
\end{table*}

\smallskip
\noindent \textit{\textbf{Aff-Wild database}} \hspace{0.2cm} 
Back in 2017, there existed some databases for dimensional emotion recognition. However, they were captured in laboratory settings and not in-the-wild. This urged us to collect the first large scale in-the-wild database and annotate it in terms of valence and arousal. To do so, we capitalized on the abundance of data available in video-sharing websites, such as YouTube and selected videos that display the affective behavior of people.  

To this end we  collected 298 videos displaying reactions of 200 subjects, with a total video duration of more than 30 hours, consisting of 1,224,100 total number of frames. This database has been annotated by 8
lay experts in a per-frame basis with regard to valence and arousal; the annotated values were continuous and ranged in [-1,1]. 
We organised the Aff-Wild Challenge based on the Aff-Wild database, in conjunction with CVPR 2017. 
Aff-Wild was developed as the first  large in-the-wild database - with a big variety of: (1) emotional states, (2) rapid emotional changes, (3) ethnicities, (4) head poses, (5) illumination conditions and (6) occlusions - used for affect recognition. 

\smallskip
\noindent \textit{\textbf{Aff-Wild2 database}} \hspace{0.2cm} 
Up to the present, there was no database that contain annotations for all main behavior tasks (valence-arousal estimation, action unit detection, expression classification). Most of the existing databases contain annotations for only one task (AffectNet is the exception, that contains annotations for two tasks). Also the existing corpora have a number of other limitations; just to name a few: non in-the-wild nature; small total number of annotations (making it impossible to train deep neural networks and generalise to other databases);  automatic or semi-automatic annotation (which is error prone and makes the annotations noisy); small number of expert annotators (making the annotations biased). 

This urged and led us to create the Aff-Wild2 database; the first and only database annotated in terms of valence and arousal, action units and expressions. Aff-Wild2 is a significant extension of Aff-Wild, through augmentation  with 260 more YouTube videosof a total duration of 13 hours and 5 minutes. The new videos have wide range in subjects': age (from babies to elderly people); ethnicity (caucasian/hispanic/latino/asian/black/african american); profession (e.g. actors, athletes, politicians, journalists); head pose; illumination conditions; occlusions; emotions. 

In total, Aff-Wild2 consists of 558 videos of 458 subjects, with  around 2,800,000 frames, showing both  subtle and extreme  human behaviors in  real-world settings. Four experts annotated the database in terms of valence and arousal; three very experienced annotators annotated 63 videos, with 398,835 frames in terms of AUs 1,2,4,6,12,15,20,25; seven experts annotated 539 videos consisting of 2,595,572 frames in terms of the 7 basic expressions. All annotations have been performed in a frame-by-frame basis. Finally, we organised the Affective Behavior Analysis in-the-wild (ABAW) Competition that utilised the Aff-Wild2 database, in conjunction with IEEE International Conference on Automatic Face and Gesture Recognition (FG) 2020.

\section{Intelligent Signal Pre-Processing \& Performance Evaluation Metrics} \label{pre-processing}

Data pre-processing consists of the steps required for facilitating  extraction of meaningful features from the data. For the visual modality, the usual steps are face detection and alignment, image resizing and image normalization. The pre-processing steps described below have been utilized in all developments presented in this paper. Let us note that no data augmentation, either on-the-fly \cite{kollias2018photorealistic,kollias2020deep2,kollias2020va} or off-the-fly has been performed.

The SSH detector \cite{najibi2017ssh} based on ResNet 
was used to extract face bounding boxes in all images.
For face alignment, at first we extracted facial landmarks. Facial landmarks are defined as distinctive face locations, such as corners of the eyes, centre of the bottom lip, tip of the nose. If they are aggregated in sufficient numbers, they can effectively describe the face shape.
In our implementations, we used the facial landmark detector in the dlib library to locate 68 facial landmarks in all frames.
We focused on  5 of them -  corresponding to the location of the left eye, right eye, nose and mouth in a prototypical frontal face -as  rigid, anchor points. Then, for every frame, we extracted the respective 5 facial landmarks and computed the affinity transformation  between the coordinates of these 5 landmarks and the coordinates of the 5 landmarks of the frontal face; we imposed next this transformation to the whole new frame to perform the alignment.
All cropped and aligned images were then resized to $96 \times 96 \times 3$ pixel resolution and their intensity values were normalized to the range $[-1, 1]$.\

In some experiments, next in this paper, the audio modality was also utilized. In these cases the audio  (mono) signal was sampled at $44,100$Hz. Then spectrograms were extracted; spectrogram frames were computed over a $33$ms window with $11$ms overlap. The resulting intensity values were normalized in $[-1,1]$ to be consistent with the visual modality.

Regarding evaluation, the metric used for measuring models' performance on valence and arousal estimation is the Concordance Correlation Coefficient (CCC). CCC evaluates the agreement between two time series (e.g., annotations and predictions) by scaling their correlation coefficient with their mean square difference. In this way, predictions that are well correlated with the annotations, but shifted in value, are penalized in proportion to the deviation. CCC takes values in the range $[-1,1]$, where $+1$ indicates perfect concordance and $-1$ denotes perfect discordance. The highest the value of the CCC the better the fit between annotations and predictions, and therefore high values are desired.
CCC is defined as follows:

\begin{equation} \label{eq:11}
\rho_c = \frac{2 s_{xy}}{s_x^2 + s_y^2 + (\bar{x} - \bar{y})^2} = \frac{2s_x  s_y \rho_{xy}}{s_x^2 + s_y^2 + (\bar{x} - \bar{y})^2},
\end{equation}

\noindent
where $\rho_{xy}$ is the Pearson Correlation Coefficient, $s_x$ and $s_y$ are the variances of all valence/arousal annotations and predicted values, $\bar{x}$ and $\bar{y}$ are the corresponding mean values  and $s_{xy}$ is the corresponding covariance value.

The evaluation metric used for measuring models’ performance on 7 basic expression classification is either the F1 Score (harmonic mean of precision and recall), or the mean diagonal value of the confusion matrix. The evaluation metric used for measuring models’ performance on action unit  detection is either the F1 Score, or an average of the F1 Score and the Accuracy metrics.

\section{Uni-task affect analysis: Dimensional Affect Recognition} \label{uni-task}

In this Section we treat affect recognition as Uni-task,  i.e., with reference to the adopted emotion model. In particular, we adopt the dimensional emotion model. Adopting the expression, or the AU model, can be similarly treated \cite{kollias2018training, kollias2019deep}.

\medskip
\medskip

\noindent \textbf{A. Single-Component Architectures: CNN plus RNN network}

\medskip
\medskip

By utilising the Aff-Wild database, we developed the AffWildNet network that successfully captured the dynamics and the in-the-wild nature of the database, providing the best performance over Aff-Wild.

At first, AffWildNet is a CNN-RNN network. The CNN part is based on the VGG-FACE, or ResNet-50 network's convolutional and pooling layers. Since low- and middle-level facial features are common in both face and facial affect recognition, we adopted the VGG-FACE network,  pre-trained with a large human faces' dataset for face recognition.
The outputs of the last pooling layer of this  CNN part, concatenated with the facial landmarks, are fed to a Fully Connected (FC) layer with 4096 or 1500 hidden units (depending on whether VGG-FACE or ResNet-50 is used). This FC layer has the role to map its two types of inputs to the same feature space, before forwarding them to the RNN part. The facial landmarks, which are provided as additional input to the network, in this way, are able to  boost the performance of the main CNN part of our model. 
The output of the FC layer is then fed to the RNN part.
The RNN is used in order to model the contextual information in the data, taking into account the temporal variations. The RNN is a 2-layered GRU with 128 units in each layer; the first layer processes the FC layer outputs; the second layer is followed by the output layer that gives the final estimates of valence and arousal. GRU units have been chosen instead of LSTM ones as they are less complex, more efficient and - as shown in the experimental evaluation - provide best results.

Three novel characteristics of  AffWildNet mainly contribute to its achieving state-of-the-art  performance. First,  the fusion of facial landmarks and features extracted from the CNN part, provides the higher part of the network with the ability to fuse the extracted by the network representations with these robust 'anchor' features. Next, the adopted loss function was based on the powerful Concordance Correlation Coefficient (CCC), which had also been the main evaluation criterion used in the Aff-Wild Challenge. In particular, the defined loss function was:

\begin{equation} \label{eq:3}
\mathcal{L}_{total} = 1 - 0.5 \times (\rho_a + \rho_v),
\end{equation}

\noindent
where $\rho_a$ and $\rho_v$ is the CCC for arousal and valence, respectively.

Furthermore, the AffWildNet was trained as an end-to-end architecture, by jointly training its CNN and RNN parts; in contrast to previous works which used to either train separately the CNN and the RNN parts, or used fixed pre-trained weights for the CNN and trained only the RNN part.

\noindent \textit{Training implementation details:}
\hspace{0.1cm} Adam optimizer; batch size of 4 and sequence length of 80; initial learning rate of 0.0001, exponentially decaying after 10 epochs; dropout probability value of 0.5.
\medskip

\noindent \textit{AffWildNet Performance Evaluation} \label{cnn_dev} 

Next, we illustrate the performance of AffWildNet, comparing it to the performance of other CNN and standard CNN-RNN architectures, as well as  to the winner of the Aff-Wild Challenge.
For the CNN architectures, we consider the ResNet-50, VGG-16 and VGG-FACE networks.
For the CNN-RNN architectures, we consider a VGG-FACE-LSTM, in which the LSTM is a 2-layered RNN fed with the outputs of the first fully connected layer of VGG-FACE.

Table \ref{table1} summarizes the CCC and MSE values obtained when applying all the above architectures, to the Aff-Wild test set. 
It shows the improvement in CCC and MSE values obtained when using the AffWildNet compared to all other developed architectures. This improvement clearly indicates the ability of the AffWildNet to better capture the dynamics in Aff-Wild. Table \ref{table1} also compares the performance of AffWildNet to that of the FATAUVA-NET, which was the winner of the Aff-Wild Challenge. AffWildNet outperformed this network, as well.

\begin{table*}[h]
\caption{CCC and MSE based evaluation of valence \& arousal predictions provided by AffWildNet, FATAUVA-Net (the winner of Aff-Wild Challange) and other state-of-the-art networks. A higher CCC and a lower MSE value indicate a better performance.}

\label{table1}
\centering
\scalebox{1}{
\begin{tabular}{ |c||c|c|c|c|c|c| }
 \hline
 \multicolumn{1}{|c||}{} & \multicolumn{3}{c|}{CCC} & \multicolumn{3}{c|}{MSE}  \\
  \hline
  & Valence & Arousal & Mean Value & Valence & Arousal & Mean Value\\
 \hline
 FATAUVA-Net & 0.40 & 0.28 & 0.34 & 0.12 & 0.10 & 0.11 \\
 \hline
 VGG-16 &0.40 &0.30 & 0.35 & 0.13 & 0.11 & 0.12  \\
 \hline
 ResNet-50 &0.43  &0.30 & 0.37 & 0.11 & 0.11 & 0.11 \\
 \hline
VGG-FACE & 0.51 & 0.33 & 0.42  & 0.10 & 0.08 & 0.09  \\
 \hline
VGG-FACE-LSTM & 0.52  &0.38 & 0.45 & 0.10 & 0.09 & 0.10  \\
\hline
 \textbf{AffWildNet} & \textbf{0.57}&  \textbf{0.43} & \textbf{0.50} & \textbf{0.08}&  \textbf{0.06} & \textbf{0.07}       \\
 \hline
\end{tabular}
}
\end{table*}

Additionally, it can be shown that  AffWildNet successfully generalizes its knowledge when used in other emotion recognition datasets and contexts. By learning complex and emotionally rich features of the AffWild database, AffWildNet constitutes a robust prior for both dimensional and categorical emotion recognition. It is the first time that such a state-of-the-art performance has been achieved. We refer the interested reader to \cite{kollias2019deep}.

\medskip
\medskip
\medskip
\medskip

\noindent \textbf{B. Multi-Component Architectures: CNN plus Multi-RNN networks}

\medskip
\medskip

Next we present novel multi-component architectures that are able to achieve excellent performance in emotion recognition.  
This is illustrated through dimensional affect analysis over the One-Minute-Gradual  Emotion (OMG-Emotion) Dataset, when using visual information. It should be mentioned that the submissions we made to the OMG-Emotion Challenge were ranked at second position for valence estimation \cite{kollias2018multi,kollias2019exploiting}.

In general, features extracted from CNN lower layers contain rich, complete and time varying information, whilst, high-level features extracted from CNN higher layers, are more specific and characteristic of the studied problem.  Taking this into account, we have developed CNN plus Multi-RNN networks; these networks extract low-, mid-  and high- level features from different layers of the CNN and feed them as inputs to multiple RNNs. The best performing networks fall into two different types, based on the adopted methodology: the first, referred as CNN-3RNN, feeds the features extracted from three CNN layers to three respective RNN subnets, whereas the other, referred as CNN-1RNN, concatenates these features and processes them  through a single RNN net. An ensemble methodology, involving both types of networks is also described.

\smallskip

\noindent \textit{1) CNN-3RNN networks}\label{cnn_3rnn1_net}

\noindent The CNN-3RNN networks include, first, the convolutional and pooling layers of VGG-FACE, followed by a fully connected layer of 4096 units. The 68 facial landmarks are concatenated with the features extracted from the last pooling layer of VGG-FACE and are fed to this FC layer. Low-, mid-  and high-level feature sets are extracted from this network and each set is fed to a 2-layer RNN (GRU) network that provides an estimate of the targeted valence and arousal values. Each RNN layer comprises 128 GRU units. 
The CNN-3RNN networks are provided with an input sequence of frames (and the corresponding landmarks of each frame) and predict, for each frame, estimates of the valence-arousal values. Their median constitute the final estimates.

\begin{figure}[htb]
\centering
\adjincludegraphics[height=18.5cm]{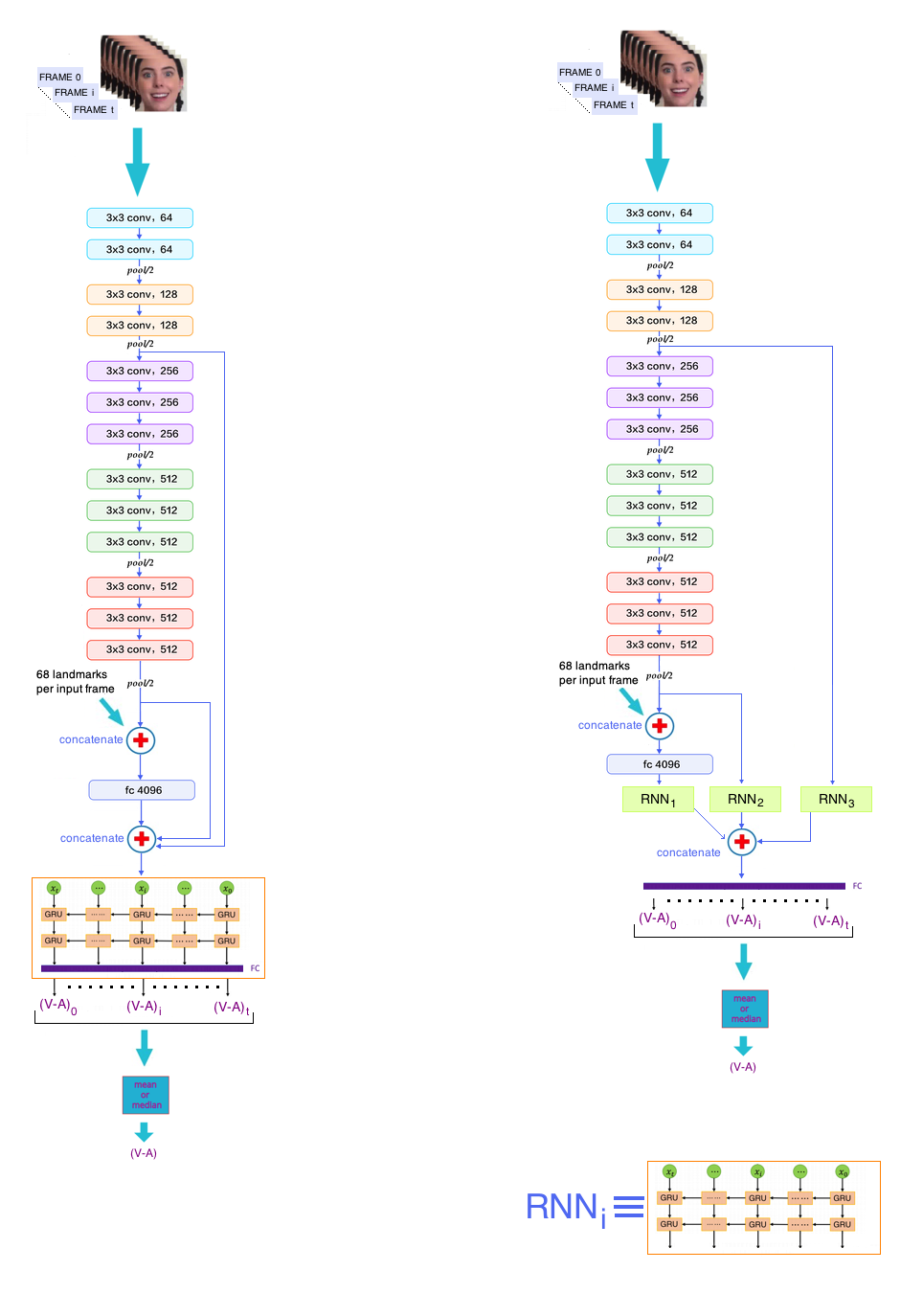}
\caption{The CNN-1RNN-2nd-pool\_last-pool\_fc architecture (on the left hand side). The CNN-3RNN-2nd-pool\_last-pool\_fc (on the right hand side). The latter architecture provided the best results. Each architecture provides a valence-arousal (V-A) estimate per input sequence of consecutive frames. The '68 landmarks' are concatenated with the features of the last 'pool' layer and passed as input to the 'fc' layer.}
\label{tac}
\end{figure}

Fig.\ref{tac} (on the right hand side) presents such a CNN-3RNN network, named CNN-3RNN-2nd-pool\_last-pool\_fc. In this network: i) the features extracted from  the FC layer are fed, as input, to a RNN network, denoted $RNN_1$; ii) the features extracted from the last pooling layer (before being concatenated with the landmarks) are fed, as input, to a second RNN network, denoted $RNN_2$; iii) the features extracted from the second pooling layer (following the fourth convolutional layer) are fed, as input, to another RNN network, denoted $RNN_3$. Fig.\ref{tac}  depicts the exact structure of the afore-mentioned $RNN_i$, $i \in\{1,2,3\}$, networks. All networks have the same structure, i.e., a 2-layer GRU network, with each layer having 128 units. Next, the outputs of the 3 RNNs are concatenated and fed to the output layer that performs the valence-arousal prediction.

\smallskip

\noindent \textit{2) CNN-1RNN networks}\label{cnn_1rnn1_net}

\noindent The CNN-1RNN type of networks also consists of the convolutional and pooling layers of VGG-FACE, followed by a FC layer of 4096 units. The 68 facial landmarks are concatenated with the features extracted from the last pooling layer of VGG-FACE and are fed to the FC layer. Low-, mid-  and high-level features are extracted, concatenated and fed to a single 2-layer GRU  that predicts the valence and arousal values. Each GRU layer comprises 128 units. 
The CNN-1RNN networks are also provided with an input sequence of frames (and the corresponding landmarks of each frame), predicting, for each frame, the valence-arousal values; their median values are the final estimates.

Fig.\ref{tac} (on the left hand side) presents such a CNN-1RNN network, named CNN-1RNN-2nd-pool\_last-pool\_fc. In this network, the features extracted from: i) the second pooling layer (following the 4th convolutional), ii) the last pooling layer (following the 13th convolutional and before being concatenated with the landmarks) and iii) the FC layer are concatenated and fed to the RNN.

\smallskip

\noindent \textit{3) Ensemble Methodology}\label{ensemble}

\noindent Next we describe an ensemble approach which fuses the above-described  networks, either at Model-level, or at Decision-level. Model-level fusion is based on concatenating the high level features extracted by the different networks, while Decision-level fusion is based on computing a weighted average of the predictions \cite{yu2021machine} provided by the different networks. On the one hand side, Model-level fusion takes advantage of the mutual information in the data. On the other hand side, the averaging procedure in Decision-level fusion reduces variance in the ensemble regressor (thus achieving higher robustness), while preserving the relative importance of each individual model. 
\smallskip

\noindent \textit{- Model-level Fusion:} \hspace{0.01cm}
Let us consider the CNN-1RNN and CNN-3RNN networks described above. We concatenate the outputs of all the RNNs in the above networks and provide them, as input, either: i) to another single RNN layer with 128 GRU units, or ii) to a fully connected layer with 128 units; the output layer follows. We denote the resulting networks as Model-level Fusion + RNN and Model-level Fusion + FC, respectively. For each frame in the input sequence of frames, this model-level fusion network predicts the valence-arousal values and then computes their  median values as final estimates.

\noindent \textit{- Decision-level Fusion:} \hspace{0.01cm}
Let us consider the CNN-1RNN and CNN-3RNN networks. The final valence (arousal) estimate $O_{v}^{dec.-level}$ ($O_{a}^{dec.-level}$), is computed as a weighted average of the final valence (arousal) estimates,  $o_{v}^{n} (o_{a}^{n})$, of these networks; each weight is proportional to the corresponding network performance on the validation set:

\begin{equation} \label{eq_1}
{O_{i}}^{dec.-level} =  \frac{1}{\sum\limits_{n}^{}{ t_i^n}} \sum_{n}^{}{ t_i^n} \cdot {o_i^n} 
,  
\end{equation}

\noindent where $i\in\{v,a\}$ ($v$ stands for valence, $a$ for arousal), 
$t_i^n $ is equal to Concordance Correlation Coefficient (defined in eq. \ref{eq:11}), for valence or arousal, computed on the validation set, with $n$ denoting the CNN-1RNN or CNN-3RNN network.

\smallskip

\noindent 
\hspace{0.3cm}
The presented multi-component networks differ from networks that either: i) use standard CNN-RNNs in which the output of the CNN is fed to the RNN, or ii) apply ensemble methodologies, using features extracted from many CNNs (but not using features from multiple layers of the same network) and fusing them. Additionally, in model-level fusion, our approach performs fusion through an RNN instead of a typical FC layer.

In addition, we performed adaptation \cite{kollias2017adaptation} of the developed architectures to the specific OMG-Emotion dataset characteristics and in particular to the dataset's annotation at utterance level. To deal with this, we split each utterance into sequences, which were individually processed by the above architectures. The median values of the predicted valence-arousal values were first computed at sequence level. Then, the median values were averaged at utterance level so as to provide the final valence and arousal estimates. This procedure deviates from related works that uniformly (or randomly) sample a constant number of frames from each utterance, assign to each of them the annotation value of the utterance and compute the prediction per frame.

Furthermore, we should mention the pre-training of the proposed architectures with the large-scale emotionally rich Aff-Wild2 database. Other works use networks that are not pre-trained on the same task (i.e., valence-arousal estimation), but on other tasks (face recognition, object detection). The pre-training on Aff-Wild2 helped our developed architectures achieve excellent performance.

\smallskip

\noindent \textit{Training implementation details:}
\hspace{0.1cm}    Adam  optimizer; loss of eq. \ref{eq:3}; end-to-end training with a learning rate of either $10^{-4}$ or $10^{-5}$; batch size of 4 and sequence length of 80; dropout with $0.5$ probability value in fully connected layers and dropout with $0.8$ probability value in the first GRU layer of the RNNs.

Finally, for all developed architectures, a chain of post-processing steps was applied. These steps included: i) median filtering of the - per frame - predictions within each sequence and ii) smoothing of the - per utterance - predictions. Any of these post-processing steps was kept when an improvement was observed on the CCC over the validation set, and applied then, with the same configuration to the test partition.

\smallskip

\noindent \textit{\textit{Multi-Component Network Performance Evaluation}} \label{experiments}

Next, we compare the CNN plus Multi-RNN architectures to the state-of-the-art CNN network, ResNet-50, and its CNN plus RNN counterpart, ResNet-RNN, as well as the state-of-the-art methods submitted to the OMG-Emotion Challenge.
Table \ref{cnn-results} illustrates the performance of all these models.

One can note that both CNN-1RNN-2nd-pool\_last-pool\_fc and CNN-3RNN-2nd-pool\_last-pool\_fc exhibit a much improved performance (around 6\% and 9\% on average) when compared to ResNet-RNN. This validates our essence that low-level CNN features together with high-level ones provide useful information for our task. Additionally, CNN-3RNN-2nd-pool\_last-pool\_fc outperformed CNN-1RNN-2nd-pool\_last-pool\_fc showing that it is better to exploit the low- and high-level features' time variations via RNNs, independently, and then concatenate them, rather than concatenate them first and process them through the use of a single RNN. 

Table \ref{cnn-results} validates that using the ensemble methodology is better than using a single network. This is because different networks produce quite different features; fusing them exploits all these representations that include rich information. It can also be observed that Model-level fusion  has a superior performance compared to that of the Decision-level one, since the features from different networks that are concatenated, contain richer information about the raw data than the final decision. In particular, in Model-level fusion, we concatenate these features and feed them to an RNN and the whole ensemble is trained end-to-end and optimised so that the concatenation of features can provide the best overall result. 
Moreover, in Model-level fusion, a better performance is achieved when a RNN, instead of a fully connected layer, is used for the fusion. Table \ref{cnn-results} also shows that our Model-level Fusion + RNN method outperforms, on both valence and arousal estimation, the winning methodologies \cite{zheng2018multimodal} of the OMG-Emotion Challenge, which have been trained additionally with the audio modality.

Another observation is that the performance of models that only used the visual modality in arousal estimation was worse than their performance in valence estimation. This was expected because, for arousal estimation, the audio cues appear to include more discriminating capabilities than facial features in terms of correlation coefficient.

\begin{table}[ht]
\caption{CCC based evaluation of valence and arousal predictions provided by the developed CNN plus Multi-RNN architectures, the state-of-the-art CNN and CNN plus RNN architectures and the winning methodologies of the OMG-Emotion Challenge. A higher CCC value indicates a better performance.}
\label{cnn-results}
\centering
\begin{tabular}{ |c||c||c|c| }
 \hline
 \multicolumn{1}{|c||}{Methods} & \multicolumn{1}{|c||}{Modality} & \multicolumn{2}{c|}{ \begin{tabular}{@{}c@{}} CCC \end{tabular}}  \\
  \hline
 &  & Valence & Arousal   \\
 \hline
 ResNet-50 & V,A: visual & 0.359  & 0.195  \\
\hline
 ResNet-RNN & V,A: visual & 0.409  & 0.224    \\
 \hline
\begin{tabular}{@{}c@{}} Single Multi-Modal \cite{zheng2018multimodal} \end{tabular}  & V,A: audio + visual   &	0.484 & 0.345  \\  
 \hline
 \begin{tabular}{@{}c@{}} Ensemble  I \cite{zheng2018multimodal} \end{tabular}   & V,A: audio + visual   &	0.496 & 0.356   \\  
 \hline
\begin{tabular}{@{}c@{}} Ensemble  II \cite{zheng2018multimodal} \end{tabular}   & V,A: audio + visual   & 0.499  & 0.361   \\
 \hline
\hline
 \begin{tabular}{@{}c@{}} \textit{CNN-1RNN-2nd-pool\_last-pool\_fc} \end{tabular} & V,A: visual & \textit{0.449}   &\textit{0.303}         \\
 \hline
  \begin{tabular}{@{}c@{}} \textbf{CNN-3RNN-2nd-pool\_last-pool\_fc} \end{tabular} & V,A: visual & \textbf{0.472}  & \textbf{0.329}    \\
 \hline
 Decision-Level Fusion & V,A: visual & 0.501  & 0.332    \\
\hline
 \textit{Model-Level Fusion + FC} & V,A: visual & \textit{0.518} &  \textit{0.348}    \\
 \hline
 \textbf{\textit{Model-Level Fusion + RNN}} & V,A: visual & \textbf{\textit{0.535}}   &  \textbf{\textit{0.365}}    \\
 \hline
\end{tabular}
\end{table}

\section{Unified Affect Analysis} \label{mt-task}

\medskip
\medskip

\noindent \textbf{A. A Multi-Task Approach to Affect Recognition}

\medskip

As has previously be mentioned, there are three main behavior tasks: i) valence-arousal estimation, ii) action unit detection and iii) basic expression classification.
Up to the present, these three tasks have been generally tackled individually from each other, despite the fact that they are interconnected. 
Multi-task learning (MTL)  is an approach that can be used to jointly learn all three behavior analysis tasks. MTL was first studied to jointly learn parallel tasks that share a common representation and to transfer part of the knowledge - learned to solve one task - to improve learning of the other related task. Several approaches have adopted MTL for solving different problems in computer vision and machine learning. In the face analysis domain, the use of MTL is somewhat limited. In the following we develop a new MTL approach to affect recognition in-the-wild, by using Aff-Wild2 which includes annotations for all three tasks.   

\smallskip

\noindent \textit{1) MT-VGG}

\noindent At first, we developed a multi-task CNN network based on the VGG-FACE. We kept the convolutional and pooling layers of VGG-FACE, discarded its fully connected layers and added on top of them 2 fully connected layers, each containing 4096 units. A (linear) output layer followed that provided final estimates for valence and arousal; it also produced 7 basic expression logits that were passed through a softmax function to get the final 7 basic expression predictions; lastly, it produced 8 AU logits that were passed through a sigmoid function to get the final AU predictions.

\noindent \textit{2) MT-VGG-GRU}

\noindent Next we extended the above described MT-VGG so as to effectively model contextual information in the data, taking into account temporal affect variations. Thus we constructed a MT CNN-RNN network. In more detail, a 2-layer GRU with 128 units per layer was stacked on top of the
first FC layer of MT-VGG; the output layer was
on top of the GRU, being the same as in MT-VGG. The prediction for each task was pooled from the same feature space, taking advantage of the correlation between the three different tasks.

\noindent \textit{3) A/V-MT-VGG-GRU}

\begin{figure}[h]
\centering
 \includegraphics[height=6.5cm]{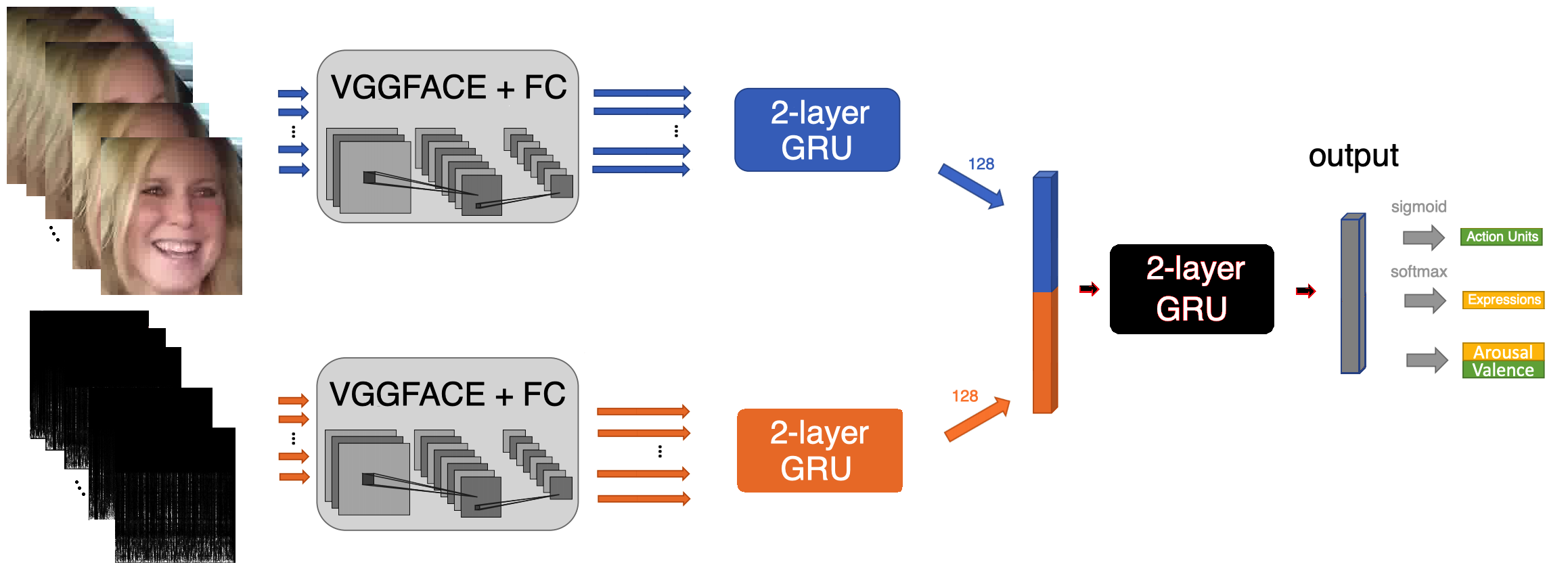} 
\caption{A/V-MT-AffWildNet: the Multi-Modal and Multi-Task developed model}
\label{multi-modal}
\end{figure}

\noindent Since Aff-Wild2 is an audiovisual (A/V) database, we additionally developed a network for handling both the video and audio modalities. 
A/V-MT-VGG-GRU consisted of two identical streams that extracted features directly from raw input images (extracted from the video) and spectrograms (extracted from the audio). Each stream consisted of a MT-VGG-GRU, without an output layer. The features from the two streams were concatenated, forming a 256-dimensional feature vector that was fed to a 2-layer GRU with 128 units in each layer, so as to fuse the information of the audio and visual streams. The output layer followed on top; it was exactly the same as in MT-VGG-GRU. A/V-MT-VGG-GRU is a multi-modal and multi-task network and is illustrated in Fig. \ref{multi-modal}. Let us note here that it is the first time  that audio is taken into account for action unit detection.

The loss function minimized during training of the multi-task networks was the sum of the individual task losses:

\begin{align}
\mathcal{L}_{CCE} &= \mathbb{E}[- \text{log } \frac{ e^{p_p}}  {\sum\nolimits_{i=1}^{7}{e^{p_i}} }] 
\label{eq:cce}\\[0.2cm]
\smallskip
\smallskip
\mathcal{L}_{BCE} &= \mathbb{E}[ -\sum\nolimits_{i=1}^{8}{(t_i \cdot \text{log } p_i + (1-t_i) \cdot \text{log } (1 -p_i))}] 
\label{eq:bce}\\[0.2cm]
\mathcal{L}_{CCC} &= 1-  0.5 \times (\rho_a + \rho_v) 
\label{eq:ccc}
\end{align}
where $\mathcal{L}_{CCE}$ is the categorical cross entropy loss, $\mathcal{L}_{BCE}$ is the binary cross entropy loss,  $p_p$  is the prediction of positive class, $p_i$ is the prediction of $AU_i$ or $Expr_i$,  $t_i$ $\in \{0,1\}$ is the label of $AU_i$, $\rho_{a,v}$ is the Concordance Correlation Coefficient (CCC) of arousal/valence.

\noindent \textit{Training implementation details:}
\hspace{0.1cm}  MT-VGG was first  pre-trained on the Aff-Wild, then the output layer was discarded and substituted by a new one for MTL. The CNN part of the MT-VGG-GRU was initialized with the weights of the MT-VGG. Training of A/V-MT-VGG-GRU was divided in two phases: first the audio/visual streams were trained independently and then the audiovisual network was trained end-to-end. To train each stream individually, we followed the
same procedure as in the MT-VGG-GRU case. Once the single streams were trained, they were used for initializing the corresponding streams in the multi-stream architecture. After pre-training, all developed networks were trained end-to-end on Aff-Wild2. We utilized  the  Adam  optimizer; learning rate was either $10^{-4}$ or $10^{-5}$; batch size was set to 10 and sequence length to 90 for MT-VGG-GRU and A/V-MT-VGG-GRU networks; batch size was set to 256 for MT-VGG; dropout with $0.5$ probability value was applied.

\smallskip

\noindent \textit{MT Approach Performance Evaluation} \label{experiments_MTL} 

Next, we compared our developed networks' performance, on a cross database experimental study, to that of state-of-the-art methods developed for these databases. Our networks were first trained on the Aff-Wild2 database and then fine-tuned to each of 9 different databases. This study illustrates that the MT networks provide the best pre-trained framework for a large variety of affect recognition settings.

Table \ref{cross_results} compares the performance, in all tasks, of the developed MT-VGG and MT-VGG-GRU (in two settings: i) when trained only with video frames; ii) when trained only with spectrograms); it also compares A/V-MT-VGG-GRU to state-of-the-art methods in Aff-Wild2, developed by the top-2 performing teams of the ABAW Competition.
It can be  observed that the MT-VGG-GRU, either when trained with the audio, or visual modality, outperformed, in all tasks, all other methods. The same happened with the best performing A/V-MT-VGG-GRU.The MT-VGG displayed a slightly worse, or similar performance in all tasks, compared to  the best performing  methods, which was expected given that these methods used either an ensemble methodology of CNN-RNNs, or fused the visual and audio modalities.

Table \ref{cross_results} also presents a cross-database  comparison on 8 databases,  between the state-of-the-art in these databases and our developed networks. Let us also note that AffectNet, RAF-DB and EmotioNet are static databases, meaning that they contain only images and thus we could only test the MT-VGG on them. AFEW-VA, DISFA, BP4DS and BP4D+ databases do not contain audio and thus we could not test the A/V-MT-VGG-GRU on them.

\begin{table}[!h]
\caption{Cross-database evaluation for the three tasks on 9 databases, between the state-of-the-art of each database and our developed networks; VA evaluation is shown as CCCV-CCCA; the mean diagonal value of the confusion matrix (denoted as 'Diag.') was the evaluation criterion for RAF-DB; '-' means that either the database did not contain audio or the database is a static one consisting of only images or the network was not trained on this database or the network was not trained for this task; $\mathcal{E}_{total}^{Expr} = 0.67 \times F_1 + 0.33 * \mathcal{T}Acc$; 
$\mathcal{E}_{total}^{AU} = 0.5 \times \mathcal{A}F_1 + 0.5 * \mathcal{T}Acc$ 
}
\label{cross_results}
\centering
\scalebox{0.73}{
\begin{tabular}{ |c||c|c|c|c|c|c|c|c|c|c|c|c| }
\hline
\multicolumn{1}{|c||}{Network} &
\multicolumn{3}{c|}{Aff-Wild2} &
\multicolumn{1}{c|}{Aff-Wild}  & \multicolumn{1}{c|}{AFEW-VA} & \multicolumn{2}{c|}{AffectNet}   & \multicolumn{1}{c|}{RAF-DB} &   \multicolumn{1}{c|}{EmotioNet}  & \multicolumn{1}{c|}{DISFA}  & \multicolumn{1}{c|}{BP4DS}  & \multicolumn{1}{c|}{BP4D+}    \\
\hline
  & CCC   & $\mathcal{E}_{total}^{Expr}$ & $\mathcal{E}_{total}^{AU}$    &  CCC   & CCC & CCC & F1 & Diag. &  F1 & F1 & F1 & F1  \\
 \hline
\hline
CNN-RNN Ensemble \cite{deng2020fau}  & 0.44-0.45  & 0.41 & 0.61 & - & - & -  & - & - & -  & - & - & -  \\
\hline
TSAV \cite{kuhnke2020two} & 0.45-0.42 &  0.51 & 0.6 & - & - & -  & - & - & -  & - & - & -  \\
\hline
AffWildNet \cite{zafeiriou,kollias2019deep} & - & - & -  & 0.57-0.43 & 0.52-0.56 & -  & - & - & -  & - & - & -  \\
\hline
AlexNet \cite{mollahosseini2017affectnet} & - & - & -  & - & - & 0.6-0.34 & \textbf{0.58} &  - & - & - & - & -  \\
\hline
VGG-FACE-mSVM \cite{li2017reliable} & - & - & -  &- & - & - & - & 0.58 & - & -  & - & -   \\
\hline
ResNet-34 \cite{ding2017facial}  &- & - & -  & - & - & -  & - & - & 0.51 & - & - & -  \\
\hline
R-T1 \cite{li2017action}& - & - & -  & - & - & - & - & - & - & 0.6 & - & -  \\
\hline
DLE extension\cite{yuce2015discriminant} & -  & - & - & -  &- & - & - & -  & - & - & 0.54 & -  \\
\hline
VGG+SVM\cite{tang2017view} & - & - & - &  - & - - & - & -  && - & - &  - & 0.51  \\
\hline
\hline
MT-VGG & 0.43-0.42 & 0.5 & 0.6 & 0.56-0.35  & 0.58-0.53 & \textbf{0.61-0.46} & \textit{0.54} & \textbf{0.61} & \textbf{0.52} & 0.61 & 0.66 & 0.49  \\
\hline
\begin{tabular}{@{}c@{}}  MT-VGG-GRU \\ (audio modality) \end{tabular} & 0.44-0.51 & 0.51 & 0.62 &0.54-0.47 & - &   - & - & - & - &- & - & - \\
\hline
\begin{tabular}{@{}c@{}} MT-VGG-GRU \\ (visual modality) \end{tabular} & 0.46-0.45 & 0.52 & 0.62 & 0.6-0.45 & \textbf{0.6-0.6} &  -  & - & - & - & \textbf{0.63} & \textbf{0.67} & \textbf{0.52}  \\
\hline
A/V-MT-VGG-GRU & \textbf{0.47-0.52} & \textbf{0.53} & \textbf{0.63} & \textbf{0.62-0.49} & - &  - & - & - & - & -  & - & -  \\
\hline
\end{tabular}
}
\end{table}

The A/V-MT-VGG-GRU achieved the best performance in Aff-Wild for both valence and arousal estimation, outperforming the existing state-of-the-art AffWildNet. 
Moreover, it can be seen that MT-VGG-GRU performed best for valence estimation when trained, on Aff-Wild, with the visual modality, whilst performed best for arousal when trained with the audio modality. This is because audio tends to have thematic constancy. Consider, for example, two fight sequences in a movie, one being a flashy fight scene and the other a one-sided fight with a person being injured. In both cases, arousal can be high due to loud and pronounced music, but valence will be positive in the former and negative in the latter sequence. 

Table \ref{cross_results} shows that the MT-VGG-GRU, trained with the visual modality, outperformed the fine-tuned AffWildNet in AFEW-VA database. It can also be observed that the MT-VGG outperformed: i) the state-of-the-art AlexNet \cite{mollahosseini2017affectnet} on AffectNet both in valence and arousal estimation, ii) the state-of-the-art VGG-FACE-mSVM \cite{li2017reliable} in RAF-DB, iii) the winner \cite{ding2017facial} of Emotionet 2017 Challenge, ResNet-34. Only, in expression recognition in AffectNet, the obtained performance of the MT-VGG is lower to the state-of-the-art. 
Finally, the MT-VGG-GRU trained with the visual modality outperformed: i) the R-TI method \cite{li2017action} in DISFA, ii) the winner of FERA 2015 Challenge, DLE extension \cite{yuce2015discriminant} and iii) the winner of FERA 2017 Challenge, VGG+SVM \cite{tang2017view}.

\medskip
\medskip

\noindent \textbf{B. A Holistic Approach to Affect Recognition in-the-wild}

\medskip
\medskip

In the previous subsection we developed  multi-task networks trained on Aff-Wild2, which are able to provide affect recognition in terms of all three emotion models. We achieved this, by exploiting the fact that Aff-Wild2 contains annotations for all three affect recognition tasks. It should be, however, mentioned that the other existing databases contain annotations for only one, or two of the tasks and not for all three of them.

In the following we present the first holistic framework for affect analysis in-the-wild, in which different emotional states, such as binary action unit activations, basic categorical emotions and continuous dimensions of valence and arousal constitute interconnected tasks that are explicable by the human's affective state. What makes it different from the approach described before is the exploration of the idea of task-relatedness, given explicitly, either from external expert knowledge, or from empirical evidence. It should be mentioned that classical multi-task literature explores feature sharing and task relatedness during training. However in such multi-task settings, one typically assumes homogeneity of the tasks, i.e. that tasks are of the same type, e.g.,  object classifiers, or attribute detectors. The main difference here is that the proposed holistic framework: (i) explores the relatedness of non-homogeneous tasks, i.e., tasks for (expression) classification, (AU) detection, (V-A) regression; (ii) operates over datasets with partial, or non-overlapping annotations of the tasks; (iii) encodes explicit relationships between tasks to improve transparency and to enable expert input. In the following: 
\begin{itemize}[leftmargin=*,noitemsep,nolistsep]
\item A flexible holistic framework is presented, which can accommodate non-homogeneous tasks, by encoding prior knowledge of task relatedness. In the experiments two effective strategies of task relatednessare evaluated: 
a) based on a cognitive and psychological study, which defines how action units are related to basic emotion categories \cite{du2014compound}, and b) inferred empirically from external dataset annotations.
\item 
An effective algorithmic approach is presented, by  coupling the tasks via co-annotation and distribution matching; its effectiveness for facial behavior analysis is illustrated with experimental studies.
\item The first holistic network for facial behavior analysis (FaceBehaviorNet) is developed, trained  end-to-end to simultaneously predict 7 basic expressions, 17 action units and continuous valence-arousal, in-the-wild. The network is trained with all publicly available in-the-wild databases that, in total, consist of over 5M images - with partial and/or non-overlapping annotations for different tasks. 
\item Experimental studies illustrate that FaceBehaviorNet greatly outperforms each of the single-task networks, validating that the network's affect recognition capabilities are enhanced when it is jointly trained for all related tasks. By further exploring feature representations learned during joint training, it is shown that a good generalisation  is achieved on the task of compound expression recognition, when no, or little, training data is available (zero-shot and few-shot learning). 
\end{itemize}

Let us start with the multi-task formulation of the facial behavior model. 
In this model we have three objectives: (1) learning seven basic emotions, 
(2) detecting activations of $17$ binary facial action units, (3) learning the intensity of the valence and arousal continuous affect dimensions. 
Our target is to train a multi-task network model to jointly achieve objectives (1)-(3). However, now we assume that for a given image $x \in \mathcal{X}$, we can have a single type of label annotations; i.e., in terms of either the seven basic emotions $y_{emo} \in \{1,2,\ldots,7\}$, or the $17$\footnote{$17$ is an aggregate of action units in all datasets; typically each dataset has from 10 to 12 AUs labelled by purposely trained annotators.} binary action unit activations $y_{au} \in \{0,1\}^{17}$, or the two continuous affect dimensions, valence and arousal, $y_{va} \in [-1,1]^{2}$.
For simplicity of presentation, we use the same notation $x$ for all images leaving the context to be explained by the label notations.  
We train the multi-task model by minimizing the following total objective, which is similar to eq. \ref{eq:cce} - \ref{eq:ccc}, with a slight change in the symbols and notations used, so as to fit the following developments:

\begin{align}
\mathcal{L}_{MT} &= \mathcal{L}_{Emo} + \lambda_{1} \mathcal{L}_{AU} + \lambda_{2} \mathcal{L}_{VA} \label{eq:mt1}\\
\mathcal{L}_{Emo} &= \mathbb{E}_{x,y_{emo}}[-\text{log } p (y_{emo}|x)]\nonumber\\ 
\mathcal{L}_{AU} &= \mathbb{E}_{x,y_{au}}[- \text{log } p (y_{au}|x)] \nonumber\\ 
&= \mathbb{E}_{x,y_{au}} \Big[-  [ \sum_{k=1}^{17} \delta_k]^{-1}  \cdot \sum_{i=1}^{17} \delta_i \cdot [y_{au}^i\text{log } p (y_{au}^i|x) + (1-y_{au}^i)\text{log } (1-p (y_{au}^i|x)) \Big]
\nonumber\\
\mathcal{L}_{VA} &= 1- 0.5 \times (\rho_a + \rho_v),\nonumber
\end{align}
where: the first term is the cross entropy loss computed over images with a basic emotion label; the second term is the binary cross entropy loss computed over images with $17$ AU activations,  $\delta_i \in \{0,1\}$ indicating whether the image contains annotation for $AU_i$; the third term measures the CCC loss  as in eq. \ref{eq:ccc}.

\smallskip

\noindent \textit{Inferring Task-Relatedness:}
\hspace{0.1cm} 
In the seminal work of \cite{du2014compound}, the authors conduct a study on the relationship between emotions (basic and compound) and facial action unit activations. The summary of the study is a Table of the emotions and their prototypical and observational action units, which we include in Table \ref{table:EmoAUs} for completeness. Prototypical action units are ones that are labelled as activated across all annotators' responses, observational are action units that are labelled as activated by a fraction of annotators. For example, in emotion \emph{happiness} the prototypical are AU12 and AU25, the observational is AU6 with weight $0.51$ (observed by 51\% of the annotators). Table \ref{table:EmoAUs} provides the relatedness between emotion categories and action units obtained from this cognitive and psychological study with human participants. 

Alternatively we inferred empirically the task relatedness from external dataset annotations. In particular, we used the Aff-Wild2 database, which is the first  in-the-wild database that contains annotations for all three behavior tasks that we are dealing with. 
At first, we trained a network for AU detection on the union of Aff-Wild2 and GFT databases \cite{girard2017sayette}. Next, this network was used to automatically annotate all Aff-Wild2 videos with AUs. Table \ref{table:EmoAUs} also shows the distribution of AUs for each basic expression. In parenthesis next to each AU is the percentage of images annotated with the specific expression in which this AU was activated.

\noindent In the following, we describe and explain the developed losses used for coupling the tasks, mentioning the case where task relatedness is inferred from the cognitive and psychological study \cite{du2014compound}.

\smallskip

\subsubsection{Coupling of basic emotions and AUs via co-annotation}

\begin{table}[t]
\caption{Relatedness between: i) basic emotions and their prototypical and observational AUs from \cite{du2014compound}: the weights $w$ in brackets correspond to the fraction of annotators that observed the AU activation; ii) basic emotions and AUs, inferred from Aff-Wild2: the weights $w$ in brackets correspond to the percentage of images annotated with the specific expression in which the AU was activated.}
\label{table:EmoAUs}
\centering
\scalebox{0.87}{
\begin{tabular}{|l|c|c|c|}
\hline
 & \multicolumn{2}{c|}{\begin{tabular}{@{}c@{}}    Cognitive-Psychological  Study \cite{du2014compound} \end{tabular}} & Empirical  Evidences, Aff-Wild2 \\
\hline
Emotion   & Prototypical AUs & Observational AUs (with weights $w$) & AUs (with weights $w$)\\
\hline\hline
happiness &  12, 25 & 6 (0.51)  & 12 (0.82), 25 (0.7), 6 (0.57), 7 (0.83), 10 (0.63) \\
\hline
sadness &  4, 15 & 1 (0.6), 6 (0.5), 11 (0.26), 17 (0.67) & (0.53), 15 (0.42), 1 (0.31), 7 (0.13), 17 (0.1) \\
\hline
fear &  1, 4, 20, 25 &2 (0.57), 5 (0.63), 26 (0.33) & 1 (0.52), 4 (0.4), 25 (0.85), 5 (0.38), 7 (0.57), 10 (0.57) \\
\hline
anger &4, 7, 24 &10 (0.26), 17 (0.52), 23 (0.29) & 4 (0.65), 7 (0.45), 25 (0.4), 10 (0.33), 9 (0.15)\\
\hline
surprise &1, 2, 25, 26 &5 (0.66)  & 1 (0.38), 2 (0.37), 25 (0.85), 26 (0.3), 5 (0.5), 7 (0.2) \\
\hline
disgust &9, 10, 17 & 4 (0.31), 24 (0.26) & 9 (0.21), 10 (0.85), 17 (0.23), 4 (0.6), 7 (0.75), 25 (0.8)\\
\hline
\end{tabular}
}
\end{table}

We propose a \emph{co-annotation} strategy to couple  training of emotions and action unit predictions. 
Given an image $x$ with the ground truth 
basic emotion $y_{emo}$, we enforce the prototypical and observational AUs of this emotion to be activated. We co-annotate the image $(x,y_{emo})$ with $y_{au}$; 
 this image contributes to both 
$\mathcal{L}_{Emo}$ and $\mathcal{L}_{AU}$\footnote{Here we overload slightly notations; for co-annotated images, $y_{au}$ has variable length and only contains prototypical/observational AUs.} in eq.  \ref{eq:mt1}. We re-weight the contributions of the observational AUs with the annotators' agreement score (from Table \ref{table:EmoAUs}).
Similarly, for an image $x$ with ground truth  
action units $y_{au}$, we check whether we can co-annotate it with an emotion label. 
{For an emotion to be present, all its prototypical and observational AUs have to be present. In cases when more than one emotion is possible, we assign the label $y_{emo}$ of the emotion with the largest requirement of prototypical and observational AUs}.
The image $(x,y_{au})$ that is co-annotated with the emotion label $y_{emo}$ 
contributes to both $\mathcal{L}_{AU}$ and $\mathcal{L}_{Emo}$ in eq. \ref{eq:mt1}. 
We use this approach to develop FaceBehaviorNet with co-annotation. 

\smallskip

\subsubsection{Coupling of basic emotions and AUs via distribution matching} 
The aim here is to align the \emph{predictions} of emotions and action units tasks during training.  
For each sample $x$ we have the predictions of emotions $p(y_{emo}|x)$ as the softmax scores over 7 basic emotions and we have the prediction of AU activations $p(y_{au}^i|x)$, $i=1,\ldots,17$ as the sigmoid scores over $17$ AUs. 
The distribution matching idea is the following: we match the distribution over AU predictions $p(y_{au}^i|x)$ with the distribution 
$q(y_{au}^i|x)$, where the AUs are modeled as a mixture over the basic emotion categories: 
\begin{equation}
    q(y_{au}^i|x) = \sum_{y_{emo} \in \{1,\ldots,7\}} p(y_{emo}|x) \: p(y_{au}^i| y_{emo}), 
\label{eq:distr}
\end{equation} 
where $p(y_{au}^i| y_{emo})$ is defined in a deterministic way from Table~\ref{table:EmoAUs} 
and is equal to 1 for prototypical/observational action units, or to 0 otherwise.  For example, AU2 is prototypical for emotion \emph{surprise} and observational for emotion \emph{fear} and thus $q(y_{\text{AU2}}|x) = p(y_{\text{surprise}}|x) + p(y_{\text{fear}}|x)$. \footnote{We also tried a variant with reweighting for observational AUs, i.e. $p(y_{au}^i| y_{emo})=w$ }.   
This matching aims to make the network's predicted AUs consistent with the prototypical and observational AUs of the network's predicted emotions. So if, e.g., the network predicts the emotion \emph{happiness} with probability 1, i.e., $p(y_{\text{happiness}}|x)=1$, then the prototypical and observational AUs of \emph{happiness}, i.e., AUs 12, 25 and 6- need to be activated in the distribution q: $q(y_{\text{AU12}}|x) = 1$; $q(y_{\text{AU25}}|x) = 1$; $q(y_{\text{AU6}}|x) = 1$; $q(y_{au}^i|x) = 0$, $i \in \{1,..,14\}$. 
In spirit of the distillation approach, we match the distributions $p(y_{au}^i|x)$ and $q(y_{au}^i|x)$ 
by minimizing the cross entropy with the soft targets loss term\footnote{This can be seen as minimizing the KL-divergence $KL(p||q)$ across the $17$ action units.}:
\begin{align}
\mathcal{L}_{DM} = \mathbb{E}_{x} \sum_{i=1}^{17}[ -p(y_{au}^i|x)\text{log }q(y_{au}^i|x) ], \label{eq:coupleloss}
\end{align}
where all available training samples are used to match the predictions.
We use this approach to develop FaceBehaviorNet with distr-matching.

\smallskip
\subsubsection{A mix of the two strategies, co-annotation and distribution matching} Given an image $x$ with the ground truth annotation of the action units $y_{au}$, we can first co-annotate it with a \emph{soft label} in form of the  distribution over emotions and then match it with the predictions of emotions $p(y_{emo}|x)$. 
More specifically, for each basic emotion, we compute the score of its prototypical and observational AUs being present. For example, for emotion \emph{happiness}, we compute $(y_{\text{AU12}} + y_{\text{AU25}} + 0.51 \cdot y_{\text{AU6}}) / (1+1+0.51)$, or set all weights to be equal to $1$, when no reweighting is used. We take a softmax over the scores to produce the probabilities over emotion categories. In this variant, every single image that has ground truth annotation of AUs will have a \emph{soft} emotion label assigned to it. Finally we match the predictions $p(y_{emo}|x)$ and the soft label by minimizing the cross entropy with soft targets similarly to eq. \ref{eq:coupleloss}. We use this approach to develop FaceBehaviorNet with soft co-annotation.

\smallskip
\subsubsection{Coupling of categorical emotions and AUs with continuous affect} 
In our work, continuous affect (valence and arousal) is implicitly coupled with the basic expressions and action units via a joint training procedure. Also one of the datasets we used has annotations for categorical and continuous emotions (AffectNet). 

\smallskip

\begin{figure}[t]
\centering
\adjincludegraphics[width=1\linewidth]{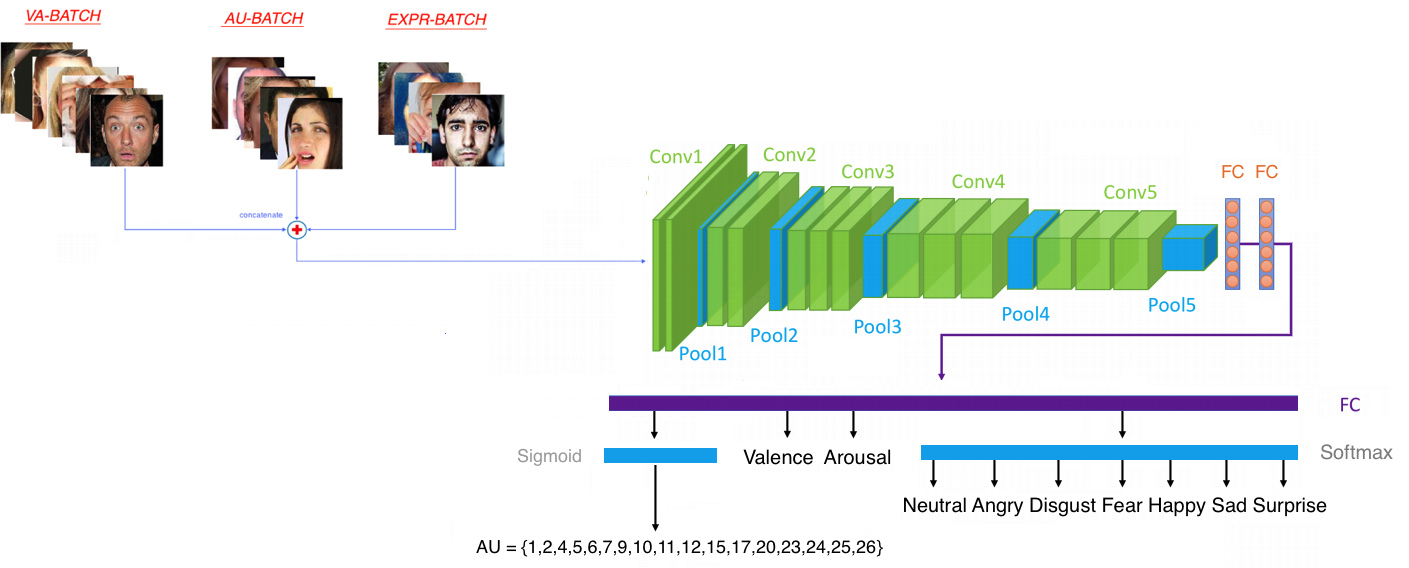} 
\caption{The holistic (multi-task, multi-domain, multi-label) FaceBehaviorNet; 'VA/AU/EXPR-BATCH' refers to batches annotated in terms of VA/AU/7 basic expressions} 
\label{facebehaviornet}
\end{figure}

\noindent \textit{FaceBehaviorNet structure \& Training implementation details:}
\hspace{0.1cm} 
Fig.\ref{facebehaviornet} shows the structure of the holistic (multi-task, multi-domain and multi-label) FaceBehaviorNet, which is an adapted version of the MT-VGG described previously. The difference is that the output layer predicts valence and arousal, 7 basic expressions and 17 action units. The predictions  for  all  tasks  are  pooled  from  the  same  feature  space. 

At this point let us describe the strategy that was used for feeding images from different databases to FaceBehaviorNet. At first, the training set was split into three different sets, each of which contained images that were annotated in terms of either valence-arousal, or action units, or seven basic expressions; let us denote these sets as VA-Set, AU-Set and EXPR-Set, respectively. During training, at each iteration, three batches, one from each set (as can be seen in Fig.\ref{facebehaviornet}), were concatenated and fed to FaceBehaviorNet. This step was important for network training, because: i) the network minimizes the objective function of eq. \ref{eq:mt1}; at each iteration, the network has seen images from all categories and thus all loss terms contribute to the objective function, ii) since the network sees an adequate number of images from all categories, the weight updates (during gradient descent) are not based on noisy gradients; this in turn prevents poor convergence behaviors; otherwise, we would need to tackle these problems, e.g. do asynchronous SGD to make the task parameter updates decoupled, 
iii) the CCC cost function needs an adequate sequence of predictions.

Since VA-Set, AU-Set and EXPR-Set had different sizes, they needed to be  'aligned'. To do so, we selected the batches of these sets in such a manner, so that after one epoch we have sampled all images in the sets. In particular, we chose batches of size 401, 247 and 103 for the VA-Set, AU-Set and EXPR-Set, respectively. 
The training of FaceBehaviorNet was performed in an end-to-end manner, with a learning rate of $10^{-4}$. A 0.5 Dropout value was used in the fully connected layers.

\smallskip

\noindent \textit{FaceBehaviorNet Performance Evaluation} \label{experiments_FBN} 

Next, we trained a VGG-FACE network on all  dimensionally annotated databases to predict valence and arousal; we also trained another VGG-FACE network on all categorically annotated databases, to perform seven basic expression classification; finally we trained a third VGG-FACE network on all databases annotated with action units, so as to perform AU detection. For brevity these three single-task networks are denoted as '(3 $\times$) VGG-FACE single-task' in one row of Table \ref{comparison_sota}.
We compared these networks' performance to the performance of FaceBehaviorNet when trained with and without the coupling losses. We also compared them to the performance of state-of-the-art methodologies developed for each utilised database, that we described previously.
Table \ref{comparison_sota} displays the performance of all these networks.

 It might be argued that the more data used for network training (even if they contain partial or non-overlapping annotations), the better network performance will be in all tasks. However this may not  be true, as the three studied tasks are non-homogeneous and each one of them contains ambiguous cases: i) there is, in general, discrepancy in the perception of the disgust, fear, sadness and (negative) surprise emotions across different people and across databases; ii) the exact valence and arousal value for a particular affect is not consistent among databases; iii) the AU annotation process is a hard to do and error prone one.
Nevertheless, from Table \ref{comparison_sota}, it can be verified that FaceBehaviorNet achieved a better performance on all databases than the independently trained VGG-FACE single-task models. This illustrates that all described facial behavior understanding tasks are coherently correlated to each other. Thus, simultaneously training an end-to-end architecture, with heterogeneous databases, leads to improved performance.

In Table \ref{comparison_sota}, it can be observed that FaceBehaviorNet trained with  no coupling loss: i) ouperforms the state-of-the-art by 3.5\% (average CCC) on Aff-Wild, 4\% (average CCC) on AffectNet, 9\% on RAF-DB and 2\% on BP4DS; ii) shows inferior performance by 4\% on AffectNet and 1\% on EmotioNet, 1\% on BP4D+. However, when FaceBehaviorNet is trained with soft co-annotation and distr-matching losses (either when task relatedness is inferred from Aff-Wild2 or from \cite{du2014compound}), it shows superior performance to all state-of-the-art methods. The fact that it outperforms these methods and the single-task networks, in both task relatedness settings, verifies the generality of the proposed losses; network performance is boosted independently of the Table of task relatedness used.

\begin{table*}[ht]
\caption{Performance evaluation of valence-arousal, seven basic expression, compound expression and action units predictions on all utilised databases provided by the FaceBehaviorNet and the state-of-the-art methods; CCC is shown as CCCV-CCCA; ’Diag.’ is the  mean  diagonal  value  of  the  confusion  matrix; 'AFA' is the Average between the mean F1 score and the mean Accuracy; 'UAR' is the Unweighted Average Recall} 
\label{comparison_sota}
\centering
\scalebox{0.77}{
\begin{tabular}{ |c||c|c|c|c|c|c|c|c|c|c|c| }
 \hline
\multicolumn{1}{|c||}{\begin{tabular}{@{}c@{}} Databases  \end{tabular}} & 
\multicolumn{1}{c|}{Aff-Wild} & \multicolumn{2}{c|}{\begin{tabular}{@{}c@{}}  AffectNet \end{tabular}} 
&  \multicolumn{2}{c|}{RAF-DB} & \multicolumn{3}{c|}{EmotioNet} & \multicolumn{1}{c|}{DISFA}  
& \multicolumn{1}{c|}{BP4DS} 
& \multicolumn{1}{c|}{BP4D+}   \\
\hline
   & VA & VA & \begin{tabular}{@{}c@{}}  Basic \\ Expr \end{tabular} & \begin{tabular}{@{}c@{}}  Basic \\ Expr \end{tabular} & \begin{tabular}{@{}c@{}}  Compound \\ Expr \end{tabular} &  \multicolumn{2}{c|}{\begin{tabular}{@{}c@{}}  Compound \\ Expr \end{tabular}} & AU & AU & AU & AU \\ 
   \hline
  & CCC & CCC &\begin{tabular}{@{}c@{}}  F1  \end{tabular} &  \begin{tabular}{@{}c@{}} Diag.   \end{tabular} & Diag. 
  
  & \begin{tabular}{@{}c@{}}  F1  \end{tabular} & UAR
  
  & AFA
  & \multicolumn{1}{c|}{\begin{tabular}{@{}c@{}}  F1  \end{tabular}}&\multicolumn{1}{c|}{\begin{tabular}{@{}c@{}}  F1  \end{tabular}} &\multicolumn{1}{c|}{\begin{tabular}{@{}c@{}}  F1  \end{tabular}}   \\ 
  \hline
 \hline
VGG-FACE\cite{zafeiriou,kollias2019deep}  & 0.51-0.33 & - & -   & - & - & - &-  & - &-  & - & - \\
\hline
AlexNet \cite{mollahosseini2017affectnet} & - &  0.60-0.34 & 0.58 & -  & - & - &-  & - &-  & - & -   \\ 
\hline 
VGG-FACE-mSVM \cite{li2017reliable}&  - &-   & - & 0.58 & 0.32 & - & -  & - &-  & - & - \\
\hline 
DLP-CNN \cite{li2017reliable} &  - &-   & - & - & 0.45 & - & - & - & - & - & - \\
\hline
NTechLab \cite{benitez2017emotionet}  & - & - & - &-  & -   & 0.26 & 0.24 & - & - & - & -  \\
 \hline
ResNet-34 \cite{ding2017facial} & - & - & - &- & -  & - & -  & 0.73 & - & - & - \\ 
\hline
DLE extension \cite{yuce2015discriminant} & - & - & -   & - & -   & -  &  - & - & - &0.59 & -  \\
\hline 
\cite{tang2017view} & - & - & - & - &-  & - & - & - & -   & -     & 0.58  \\
\hline
\hline
(3 $\times$) VGG-FACE single-task & 0.52-0.31 & 0.53-0.43 & 0.51 & 0.59 & -  & - & - &0.67 & 0.47 & 0.56 & 0.54 \\
\hhline{=:=:=:=:=:=:=:=:=:=:=:=}
\begin{tabular}{@{}c@{}}  FaceBehaviorNet \\ no coupling loss \end{tabular} & \textit{0.55-0.36} & 0.56-\textit{0.46} & 0.54  & \textit{0.67} & - & - & - & 0.72 & \textit{0.52} & \textit{0.61} & 0.57  \\
\hline
\begin{tabular}{@{}c@{}} \textbf{FaceBehaviorNet   soft co-annotation} \\ \textbf{\& distr-matching, \cite{du2014compound}}  \end{tabular} & 0.59-\textbf{0.41}  & 0.59-0.50  & \textbf{0.60}     & 0.70 & - & - & -& 0.73   & 0.57  & \textbf{0.67}  & \textbf{0.60}  \\
\hline
\begin{tabular}{@{}c@{}} \textbf{FaceBehaviorNet  soft co-annotation} \\ \textbf{ \& distr-matching, Aff-Wild2}  \end{tabular} & \textbf{0.60}-0.40  & \textbf{0.61}-\textbf{0.51}  & \textbf{0.60}      & \textbf{0.71} & - & - & -  & \textbf{0.74}   & \textbf{0.60}  & 0.66  & \textbf{0.60}  \\ 
\hline
\hline
\begin{tabular}{@{}c@{}}  zero-shot FaceBehaviorNet \\ no coupling loss \end{tabular}& - & - & - & - &    0.35 &0.25 & 0.27   & - & - & - & -   \\
\hline
\begin{tabular}{@{}c@{}} \textbf{zero-shot  FaceBehaviorNet} \\ \textbf{soft co-annotation} \\  \textbf{\& distr-matching \cite{du2014compound}}  \end{tabular} & - & - & - & -  &  \textit{0.37} & \textbf{0.32}  & \textbf{0.34}    & - & - & - & -   \\
\hline
\hline
\begin{tabular}{@{}c@{}} fine-tuned FaceBehaviorNet \\ no coupling loss \end{tabular} & - & - & - & -  &  0.46 & - & -  & - & -  & - & -    \\
\hline
\begin{tabular}{@{}c@{}} \textbf{fine-tuned FaceBehaviorNet} \\  \textbf{soft co-annotation} \\  \textbf{\& distr-matching \cite{du2014compound}} \end{tabular} & -  & - & - & -  & \textbf{0.49}  & - & - & - & - & - & -  \\
\hline
\end{tabular}
}
\end{table*}

\smallskip

\noindent \textit{Zero-Shot and Few-Shot Learning}  
\hspace{0.1cm} 
In order to further prove and validate that FaceBehaviorNet learned good features encapsulating all aspects of facial behavior, we conducted zero-shot learning experiments for classifying compound expressions.   
Given that there exist only 2 datasets (EmotioNet and RAF-DB) annotated with compound expressions and that they do not contain a lot of samples (less than 3,000 each), at first, we used the predictions of FaceBehaviorNet together with the rules from \cite{du2014compound} to generate compound emotion predictions. Additionally, to  demonstrate  the  superiority  of FaceBehaviorNet, we used it as a pre-trained network in a few-shot learning experiment. We took advantage of the fact that the network has learned good features and used them as priors for fine-tuning the network to perform compound emotion classification.

\paragraph{RAF-DB database}

At first, we performed zero-shot experiments on the 11 compound categories of RAF-DB. We computed a candidate score,  $\mathcal{C}_{s}(y_{emo})$, for each class $y_{emo}$: 



\begin{align}
 \mathcal{C}_{s}(y_{emo}) &= \frac{ \sum_{k=1}^{17} p(y_{au}^k| y_{emo})} { \sum_{k=1}^{17} p(y_{au}^k|x) \: p(y_{au}^k| y_{emo}) }
  &+ p(y_{emo1}) + p(y_{emo2}) 
 &+ 0.5 \cdot (\frac{p(y_{v}|x)}{|p(y_{v}|x)|} + 1),     p(y_{v}|x) \neq 0 
\end{align}

\noindent
where: i) the first term of the sum is
 FaceBehaviorNet's predictions of only the prototypical (and observational) AUs that are associated with this compound class according to \cite{du2014compound}; in  this  manner,  every AU  acts  as  an  indicator  for  this  particular  emotion  class; ii) $p(y_{emo1})$ and $p(y_{emo2})$ are FaceBehaviorNet's predictions of  the basic expression classes $emo1$ and $emo2$ that are mixed and form the compound class (e.g., if the compound class is happily surprised then $emo1$ is happy and $emo2$ is surprised); iii) the last term of the sum is added only to the happily surprised and happily disgusted classes and is either 0 or 1 depending on whether FaceBehaviorNet's valence prediction is negative or positive, respectively; the rationale is that only happily surprised and (maybe) happily disgusted classes have positive valence; all other classes are expected to have negative valence as they correspond to negative emotions. 
 Our final prediction was the class that had the maximum candidate score.

Table \ref{comparison_sota} shows the results of this approach when we used the predictions of FaceBehaviorNet trained with and without the soft co-annotation and distr-matching losses. Best results have been obtained when the network was trained with the coupling losses. One can observe, that this approach outperformed by 5\% the VGG-FACE-mSVM \cite{li2017reliable} which has the same architecture as this network and has been trained for compound emotion classification. 

Next, we target few-shot learning. In particular, we fine-tune the FaceBehaviorNet (trained with and without the soft co-annotation and distr-matching losses) on the small training set of RAF-DB. In Table \ref{comparison_sota} we compared its performance to a state-of-the-art network. It can be seen that the fine-tuned FaceBehaviorNet, trained with and without the coupling losses, outperformed by 1\% and 4\%, respectively, the best performing network, DLP-CNN, that was trained with a loss designed for this specific task.

\paragraph{EmotioNet database}

Next, we performed zero-shot experiments on the EmotioNet basic and compound set that was released for the related Challenge. This set includes 6 basic plus 10 compound categories. The zero-shot methodology we followed was similar to the one described above for the RAF-DB database. 
The results of this experiment can be found in Table \ref{comparison_sota}. 
Best results have also been obtained when the network was trained with the two coupling losses. It can be observed that this approach outperformed by 6\% and 10\% in F1 score and Unweighted Average Recall (UAR), respectively, the state-of-the-art NTechLab's \cite{benitez2017emotionet} approach, which used the Emotionet's images with compound annotation.

\section{Conclusions} \label{conclusion}
This paper presented the state-of-the-art in affect recognition in-the-wild. This involved large databases annotated in terms of: dimensional emotion variables,  i.e.,  valence and arousal; seven basic expression categories; facial action units. This also involved deep neural architectures \cite{kollias2018deep}  that are trained with these databases, providing state-of-the-art performance on them, or constituting robust priors for dimensional, and/or categorical recognition over other datasets in-the-wild. 

In this framework, we presented the state-of-the-art affect recognition DNN, AffWildNet, trained on the Aff-Wild database. 
We also presented a DNN methodology by extracting  low-, mid- and high-level latent information and analysing this by multiple RNN subnets. Moreover, we showed that an ensemble approach, based on model-level fusion,  produced excellent results for visual affect recognition on the OMG-Emotion Challenge.
We further explored the use of Aff-Wild2, the largest in-the-wild, audiovisual, database, being annotated in terms of valence-arousal dimensions, seven basic expressions and facial action units. We  presented  multi-task DNNs, being trained on Aff-Wild2, illustrating that they outperform the state-of-the-art for affect recognition. 
We additionally presented the FaceBehaviorNet, which is the first holistic framework for joint: basic expression recognition, action unit detection and valence-arousal estimation.

Future directions include the  development of new scalable architectures that can learn to analyse affect, by extracting coarse-to-fine information from visual inputs in-the-wild.  Moreover, focusing on specific types of affect, for example, related to negative or reluctant behaviours, or on compound emotions, will require extending the databases, and/or performing domain adaptation of the developed architectures in these frameworks.   
Unsupervised learning will be the main direction in the future for handling non-annotated data cases, while focusing on related problems, such as uncertainty of the estimation procedure. 
Moreover, although it has been shown that DNNs are capable of analysing large datasets for affect recognition, they lack transparency in their decision making, in the sense that it is not straightforward to justify their prediction. Extraction of latent variables, such as low-, medium- and high-level features and further unsupervised exploration of them in multiple contexts and in low-shot learning frameworks can be further investigated.


%

\bibliographystyle{IEEEtran}
\bibliography{main}

%

\begin{IEEEbiography}[{\includegraphics[width=1in,height=1.25in,clip,keepaspectratio]{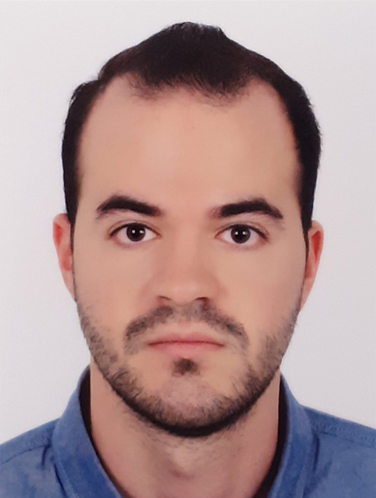}}]{Dimitrios Kollias}, Fellow of the Higher Education Academy, holder of a Post-Graduate Certificate and member of the IEEE, is currently a Senior Lecturer in Computer Science with the School of Computing and Mathematical Sciences, University of Greenwich. He has been the recipient of the prestigious Teaching Fellowship of Imperial College London. He has obtained the Ph.D. from the Department of Computing, Imperial College London, where he was a member of the iBUG group. Prior to this, he received the Diploma/M.Sc. in Electrical and Computer Engineering from the ECE School of the National Technical University of Athens, Greece, and the M.Sc. in Advanced Computing from the Department of Computing of Imperial College London. He has published his research in the top journals and conferences on machine learning, perception and computer vision such as IJCV, CVPR, ECCV, BMVC, IJCNN, ECAI and SSCI. He is a reviewer in top journals and conferences, such as CVPR, ECCV, ICCV, AAAI, TIP, TNNL, TAC, Neurocomputing, Pattern Recognition and Neural Networks.
He has been Competition Chair and Workshop Chair in IEEE FG 2020. He has won many grants and awards, such as from the City and Guilds College Association, the Imperial College Trust and the Complex \& Intelligent Systems Journal. He has h-index 17 and i10-index 19.  
His research interests span the areas of machine and deep learning, deep neural networks, computer vision, affective computing and medical imaging. 
\end{IEEEbiography}

\begin{IEEEbiography}[{\includegraphics[width=1in,height=1.25in,clip,keepaspectratio]{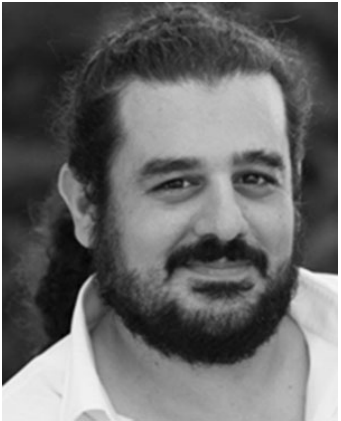}}]{Stefanos Zafeiriou} is currently a Professor in Machine Learning and Computer Vision with the Department of Computing, Imperial College London. He also holds an EPSRC Fellowship. He received the Prestigious Junior Research Fellowships from Imperial College London in 2011 to start his own independent research group. He received the President’s Medal for Excellence in Research Supervision for 2016. He received various awards during his doctoral and postdoctoral studies. He has been a Guest Editor of more than 6 journal special issues and co-organized more than 15 workshops/special sessions on specialized computer vision topics in top venues, such as CVPR/FG/ICCV/ECCV (including three very successfully challenges run in ICCV’13, ICCV’15 and CVPR'17 on facial landmark localisation/tracking). He has coauthored more than 70 journal papers mainly on novel statistical machine learning methodologies applied to computer vision problems, such as 2-D/3-D face analysis, deformable object fitting and
tracking, shape from shading, and human behavior analysis, published in the
most prestigious journals in his field of research, such as TPAMI,
IJCV, TIP, TNNLS and many papers in top conferences, such
as CVPR, ICCV, ECCV, ICML. His students are frequent recipients of very
prestigious and highly competitive fellowships, such as the Google, Intel and Qualcomm ones. He has more than
12000 citations to his work, h-index 54, i10-index	159. He was the General Chair of BMVC 2017.
\end{IEEEbiography}


\end{document}